\documentclass{trbunofficial}
\usepackage{graphicx}
\usepackage{threeparttable}
\usepackage{multirow}
\usepackage{threeparttable}
\usepackage{caption}
\usepackage{hhline}
\usepackage{pifont}
\usepackage{graphicx}
\usepackage{makecell}
\usepackage{textcase}
\usepackage{color,soul}
\soulregister\cite7
\soulregister\ref7
\usepackage[hidelinks]{hyperref}

\AuthorHeaders{Zhang, Sun, Wang, Nie, Ma, Li, Sun and Ban}

\title{\NoCaseChange{Large Language Models for Mobility Analysis in Transportation Systems: A Survey on Forecasting Tasks}
}

\author{%
  \textbf{Zijian Zhang, Co-first Author}\\
  The Department of Computer and Information Science\\
  University of Pennsylvania, Philadelphia, Pennsylvania, 19104\\
  zzjharry@alumni.upenn.edu\\
  \hfill\break%
  \textbf{Yujie Sun, Co-first Author}\\
  Philadelphia, Pennsylvania, 19104\\ 
  yjshiki@alumni.upenn.edu\\
  \hfill\break%
  \textbf{Zepu Wang, Co-first Author, Corresponding Author}\\
  Civil and Environmental Engineering Department\\
  University of Washington, Seattle, Washington, 98195\\
  zepu@uw.edu\\
  \hfill\break%
  \textbf{Yuqi Nie}\\
  Department of Electrical and Computer Engineering\\
  Princeton University, Princeton, New Jersey, 08540\\
  ynie@princeton.edu\\
  \hfill\break%
  \textbf{Xiaobo Ma}\\
  Department of Civil and Architectural Engineering and Mechanics\\
  The University of Arizona, Tucson, Arizona, 85721\\
  xiaoboma@arizona.edu\\
  \hfill\break%
  \textbf{Ruolin Li}\\
  Department of Civil and Environmental Engineering\\
  University of South California, Los Angeles, California, 90089\\
  ruolinl@usc.edu\\
  \hfill\break%
  \textbf{Peng Sun, Corresponding Author}\\
  Duke Kunshan University\\
  peng.sun568@dukekunshan.edu.cn\\
  \hfill\break%
  \textbf{Xuegang Ban}\\
  Civil and Environmental Engineering Department\\
  University of Washington, Seattle, Washington, 98195\\
  banx@uw.edu
}

\WordsPerTable{250}

\TotalWords{6974}

\begin{document}
\maketitle

\section{Abstract}

Mobility analysis is a crucial element in the research area of transportation systems. Forecasting traffic information offers a viable solution to address the conflict between increasing transportation demands and the limitations of transportation infrastructure. Predicting human travel is significant in aiding various transportation and urban management tasks, such as taxi dispatch and urban planning. Machine learning and deep learning methods are favored for their flexibility and accuracy. Nowadays, with the advent of large language models (LLMs), many researchers have combined these models with previous techniques or applied LLMs to directly predict future traffic information and human travel behaviors. However, there is a lack of comprehensive studies on how LLMs can contribute to this field. This survey explores existing approaches using LLMs for time series forecasting problems for mobility in transportation systems. 
We provide a literature review concerning the forecasting applications within transportation systems, elucidating how researchers utilize LLMs, showcasing recent state-of-the-art advancements, and identifying the challenges that must be overcome to fully leverage LLMs in this domain.

\hfill\break%
\noindent\textit{Keywords}: Large Language Models, Transportation Systems, Forecasting, Mobility, Deep Learning, Traffic Prediction
\newpage

\section{Introduction}

\begin{figure*}[!h]
\centering
\includegraphics[width=0.99\columnwidth]{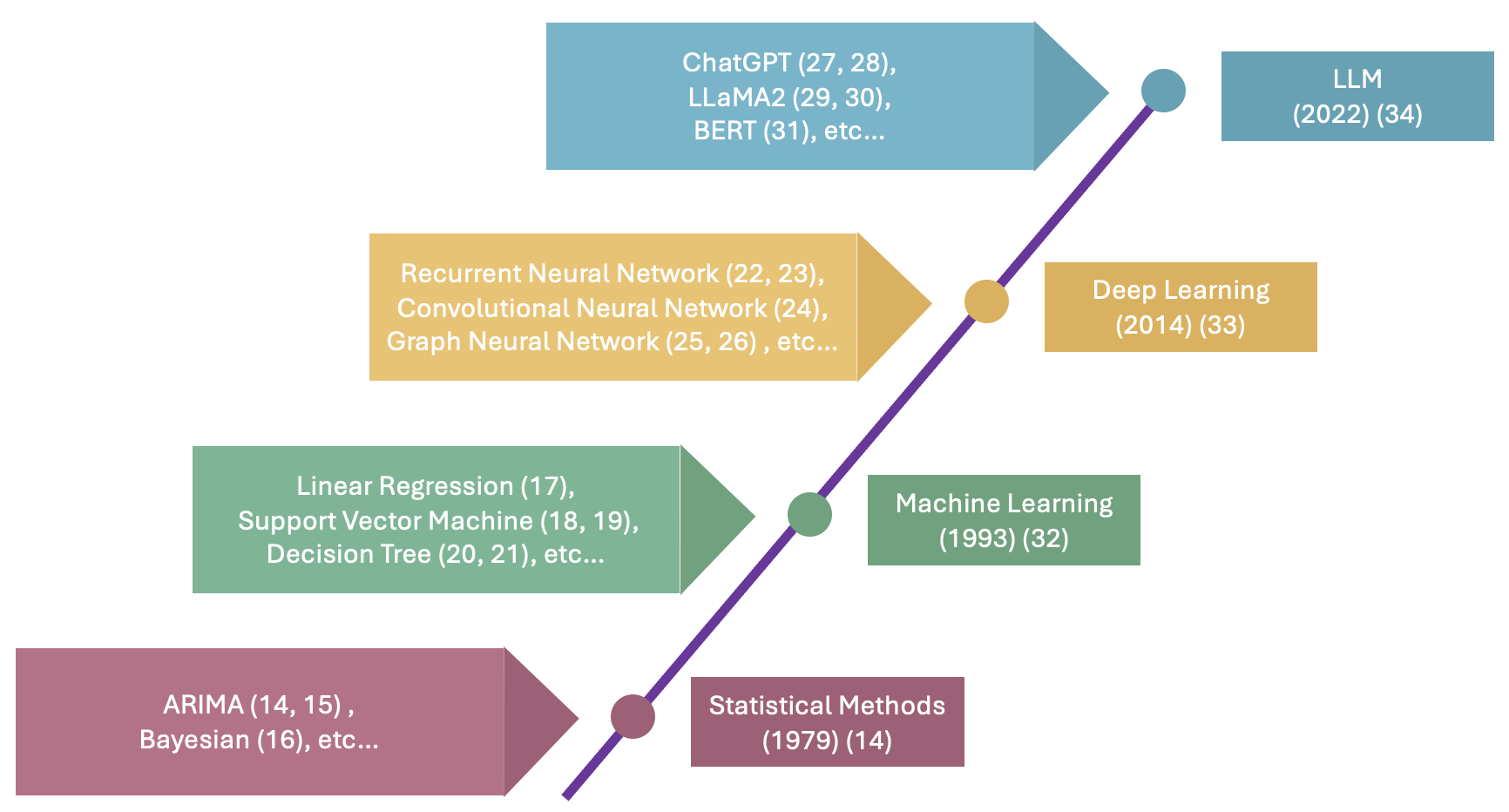}
\caption{Traffic Forcasting Technology Development \cite{ahmed1979analysis,azari2019cellular,wang2014new,sun2002short,zhang2009traffic,tang2019traffic,alajali2018intersection, xia2017traffic,dalgkitsis2018traffic,li2017diffusion,yu2016data,jiang2023graph,chen2003traffic,haydari2024mobilitygpt,zhang2023spatio,liu2024spatial,guo2024explainable,zhang2024semantic,yu1993traffic,lv2014traffic,xue2022leveraging}  
} 


 \vspace{-0.3cm}
\label{fig:timeline}
\end{figure*}

Forecasting the mobility of vehicles and pedestrians is crucial for planning and optimizing transportation systems that enable the movement of people and goods within and across different areas \cite{ ghalehkhondabi2019review, wang2022novel,wang2023sst1}. Traditionally, statistical models have been widely used for transportation system forecasting, focusing on factors such as population growth, urban development, and changes in infrastructure. Common methods include the ARIMA model, Bayesian approaches, and others.
Recently, there has been a notable shift toward leveraging deep learning techniques in this domain \cite{mamede2023deep, nie2022time, wang2023novel3, wang2024sk}. Deep learning models are extensively employed in modern scientific research and engineering \cite{zhang2021form, mulvey2022applications,long2022deep}. They particularly excel at identifying complex patterns in mobility data, offering insights into traffic flow and public transit demand with high accuracy \cite{ma2024data, wang2023stmlp,utku2023new}. The developmental timeline of these advancements is illustrated in Figure \ref{fig:timeline}.

The development of Large Language Models (LLMs) has introduced a new paradigm for quantitative problem solving in various domains \cite{topsakal2023creating, leong2024metroberta, nie2024survey}. 
These models, exemplified by the Generative Pre-trained Transformer (GPT) series, have significantly impacted research areas ranging from sentiment analysis, machine translation, and text summarization in Natural Language Processing (NLP), as well as data augmentation, predictive modeling, big data analytics, and statistical learning to complex data analysis \cite{yenduri2024gpt, imamguluyev2023rise, nazir2023comprehensive}. 
LLMs stand out for their ability to process and interpret large datasets in a sophisticated manner, closely mirroring human cognitive abilities \cite{hagendorff2023human}. This capability makes them particularly promising for applications in understanding diverse and complex data streams \cite{ma2023eliminating,cottam2024large,ma2020statistical}.

Recently, the application of LLMs in time series forecasting has garnered increasing attention and progress \cite{jiang2024empowering, jin2024position, liang2024foundation}. Two primary approaches have emerged in this domain: First, researchers have developed specialized time series foundation models inspired by LLM architectures \cite{rasul2023lag, shi2024time}, as well as multi-modal foundation models capable of time series analysis \cite{zhang2023meta}. Second, investigators have explored the adaptation of pre-trained LLMs for time series forecasting through various methods, including fine-tuning \cite{zhou2023one}, reprogramming \cite{jin2024time}, and zero-shot inference \cite{gruver2024large}. LLMs distinguish themselves from traditional methods by their advanced reasoning and contextual understanding capabilities, which allows for deciphering complex patterns in data, and their flexibility in transfer learning, which minimizes the need for retraining, especially when the downstream data size is limited. Moreover, their scalability makes them suitable for real-time analysis, and their ability to handle multi-modal data is invaluable for integrating diverse data sources. LLMs also offer the potential for enhanced interpretability and customization, which are essential for practical applications where understanding the model's reasoning is crucial. These capabilities collectively highlight the potential of LLMs to revolutionize complex, multi-modal forecasting tasks in various real-world settings.

In transportation systems, time series forecasting represents a fundamental analytical task that often requires processing temporal data alongside diverse contextual information. The multi-modal nature of transportation data—encompassing structured temporal sequences (e.g., traffic flow, speed, accident and occupancy data) and unstructured contextual information—presents an ideal use case for LLM applications \cite{de2023llm, shoaib2023survey}. This contextual information may include real-time traffic incident reports, regulatory notifications from transportation authorities, visual data from traffic surveillance systems, and meteorological conditions affecting road networks. The inherent capability of LLMs to process and synthesize diverse data types while maintaining temporal coherence makes them particularly suitable for transportation forecasting tasks. For instance, LLMs can simultaneously analyze historical traffic patterns while incorporating relevant external factors such as scheduled events, weather forecasts, or infrastructure maintenance schedules, which is a task that traditionally required multiple specialized models \cite{zhang2024advancing}. Furthermore, the sophisticated pattern recognition and transfer learning capabilities of LLMs suggest their potential to address common challenges in transportation forecasting, such as handling non-linear relationships, accounting for seasonal variations, and adapting to evolving urban mobility patterns. The natural language processing capabilities of LLMs also offer the possibility of generating interpretable forecasts accompanied by contextual explanations, which could significantly enhance decision-making processes in transportation management systems \cite{guo2024explainable, peng2024lc}.

However, the specific application of LLMs in time series forecasting in transportation and urban systems has not been thoroughly explored in the existing literature. While there are studies on LLM applications in time series analysis 
\cite{jia2024gpt4mts, jin2024time, huang2024large,dan2024image} and deep learning's broader impact on transportation \cite{mamede2023deep, jiang2023graph}, a focused examination of LLMs in this context is missing. This gap indicates a significant opportunity for in-depth research on the use of LLMs for advanced traffic predictions and transportation infrastructure planning.

Our survey seeks to address this gap by presenting a comprehensive exploration of the potential of LLMs in forecasting tasks in transportation systems. We will discuss two key sets of techniques—data processing and model framework—that demonstrate the versatile applications of LLMs in both transportation and human mobility forecasting contexts. Through reviewing current research and practical applications, our work aims to highlight the transformative potentials that LLMs offer to improve the efficiency, safety, and sustainability of transportation systems, while also generating transportation and mobility planning solutions. By contributing a concentrated analysis on the role of LLMs in transportation and human mobility forecasting, we aspire to stimulate further research and innovation in this domain, as well as facilitate a richer integration of LLMs with transportation systems and human mobility planning strategies.




\section{Background} 
\label{Section: Background}

\subsection{Large Language Models (LLMs)}



In recent years, there has been a significant transformation in the field of NLP, primarily driven by the advent and evolution of LLMs. In 2018, the introduction of Bidirectional Encoder Representations from Transformers (BERT) by Devlin et al. \cite{kenton2019bert} marked a significant advancement in pre-training language representations. BERT established a new benchmark for state-of-the-art performance across a multitude of language understanding tasks by leveraging bidirectional training in a novel way. The release of GPT-3 in 2020 further expanded these capabilities by introducing and demonstrating the effectiveness of few-shot learning \cite{brown2020language}. These advancements provide a guideline on how to further improve LLM performance. Besides the models above, many new LLMs, like LLaMA \cite{touvron2023llama} and Mixtral \cite{jiang2023mistral}, are also developed, and applied to various tasks \cite{yu2023temporal,de2023llm}.

LLMs have seen diverse applications across various time series fields, including finance \cite{yu2023temporal}, healthcare \cite{peng2023study,shi2022using}, traffic management \cite{grigorev2024integrating}, and videos \cite{wang2024videoagent,xiong2024search}, demonstrating their versatility beyond traditional text-based tasks \cite{jin2024position, mulvey2022applications}. For instance, in the financial domain, researchers have leveraged LLMs to surpass conventional models like ARMA-GARCH by employing techniques such as zero-shot/few-shot inference and instruction-based fine-tuning, highlighting LLMs' capability for enhanced predictive accuracy \cite{yu2023temporal}. In healthcare, innovations like GatorTronGPT focus on medical research, including biomedical natural language processing, showcasing the potential of LLMs in processing and interpreting complex medical data \cite{peng2023study}.

The application of LLMs to traffic problems exemplifies their ability to analyze and forecast time series patterns in mobility and transportation data, further underscoring the transformative impact of LLMs across diverse research areas and practical applications.

\subsection{Forecasting Tasks in Mobility Analysis}

Time series prediction is a vital component of intelligent transportation systems due to its ability to provide predictive and timely information that benefits society at large. Normally, mobility forecasting tasks are often categorized as a type of time series prediction problem \cite{rahman2023deep,li2024demand} and deep learning techniques are the most popular approaches today. In this survey, we mainly focus on four types of time series forecasting problems: traffic forecasting, human mobility forecasting, demand forecasting and missing data imputation.


\subsubsection{Traffic Forecasting}

Traffic forecasting involves predicting future traffic conditions, such as vehicle flow, speed, and congestion levels, on transportation networks. Accurate traffic forecasts are essential for effective traffic management, infrastructure planning, and mitigating congestion in intelligent transportation systems.

The traditional approaches for traffic forecasting are usually based on time series analysis. These methods model traffic data as time-dependent sequences to identify patterns and make future predictions. The Auto-Regressive Integrated Moving Average (ARIMA) model is a widely used technique in this category. And there have been comprehensive studies on applying ARIMA models to forecast short-term traffic flow, demonstrating their effectiveness in capturing temporal dependencies in traffic data \cite{yu2004switching, alghamdi2019forecasting}.

Recently, the machine learning based models, especially deep learning methods, become more popular in the field of traffic forecasting due to their strong performance. For instance, Recurrent Neural Networks (RNNs) and Long Short-Term Memory (LSTM) networks, are used to model complex temporal patterns in traffic data, achieving high accuracy by capturing long-term dependencies in traffic flow \cite{dalgkitsis2018traffic}. In addition, the Graph Neural Networks (GNN) approaches are well suited to traffic forecasting problems because of their ability to capture spatial dependency, and there has been comprehensive studies summarizing the GNN paradigms in the traffic domain \cite{jiang2022graph}.

\subsubsection{Human Mobility Forecasting}

Besides traffic forecasting, human mobility is another significant area of interest within transportation systems research. Human mobility includes not only the movement of individuals and crowds over time and space but also their implications on traffic flow, human travel planning, and public transportation utilization. Accurate human mobility forecasting are essential for urban planning, transportation management, and public health interventions.

Among the traditional statistical methods, Markov Chains (MC) have are popular probabilistic models predict future human locations based on the current state and transition probabilities. For instance, Lu et al. \cite{lu2013approaching} proposed and implemented a series of MC based models for human forecasting, demonstrating their effectiveness in capturing sequential movement behaviors.

Similar to traffic forecasting, deep learning methods have been widely applied in the field of human mobility forecasting \cite{luca2021survey}. For example, in T-CONV \cite{lv2018t}, the authors leveraged Convolutional Neural Networks (CNN) to model trajectories as two-dimensional
images, and adopts multi-layer CNN to combine multi-scale trajectory patterns to achieve precise prediction. In addition, Xue et al. \cite{xue2021mobtcast} proposed MobTCast, which is a transformer-based model for human mobility forecasting, leveraging auxiliary trajectory forecasting to enhance accuracy.

\subsubsection{Demand Forecasting}

Traffic demand forecasting denotes the process of predicting the size of crowds or the number of vehicles traveling in a given location at a specific time in the future.

Rule-based models are traditional approaches for demand forecasting. For example, Zhao et al. \cite{zhao2018improving} presented three such models aiming at performing traffic demand forecasting with big data. The Total Sample Demand Distribution Model uses comprehensive population data to predict travel demand across regions, eliminating the need for traditional sample surveys and parameter estimation, which are required in older gravity models. The Transportation Integration Model merges several stages of traffic forecasting—such as trip distribution, mode choice, and traffic assignment—into a unified approach, allowing for real-time data integration to more accurately predict shifts in traffic patterns and congestion. Finally, the Non-Motorized Demand Forecasting Model targets demand forecasting for non-motorized modes like walking and cycling. This model uses high-resolution spatial data to improve prediction accuracy, addressing the limitations of traditional models that often overlook or inadequately predict non-motorized travel demand.

The utilization of textual information in traffic demand forecasting has been explored in some deep learning studies. For instance, two deep learning architectures, DL-LSTM and DL-FC, were proposed in \cite{rodrigues2019combining} which improved time series forecasting accuracy by leveraging text information in addition to original time-series data. These two deep learning architectures demonstrate significantly reduced forecast errors in the context of taxi demand prediction.

\subsubsection{Missing Data Imputation}

Imputation is also a critical study in traffic data studies. Due to various reasons, such as broken devices or lack of stable measuring equipment, some pieces of data in a whole traffic system may be missing. Hence, performing traffic data imputation to recover missing data is usually a necessary task in traffic research~\cite{du2024tsi}.

In early days, popular approaches for missing data imputation include traditional traffic prediction models, interpolation-based methods, and statistical learning-based methods \cite{wu2020imputation}. Traffic prediction models like autoregressive integrated moving average (ARIMA) and Bayesian networks (BNs) predict missing data using historical information. As a result, these models cannot fully utilize data collected after the missing point. Interpolation-based methods are divided into two subgroups: temporal-neighboring methods and pattern-similar methods \cite{li2014missing}.  These methods assume that traffic patterns are highly similar, limiting their application to very stable or regular situations. Statistical learning-based methods mainly include principal component analysis-based techniques such as probabilistic principal component analysis (PPCA) (as noted by \cite{qu2009ppca}). A comparative experiment among these methods was conducted by \cite{li2014missing}. The study found that the performance of different methods varies with the missing data pattern and ratio.

In recent years, deep learning models have been applied to traffic data imputation. Generative Adversarial Networks (GANs) generate realistic data through a generator-discriminator setup, improving imputation performance. Graph Neural Networks (GNNs), particularly Convolutional GNNs (or Graph Convolutional Networks, GCNs), are widely used as they utilize convolutional neural networks to embed graph information into a tensor, creating a uniform framework that can handle irregular datasets \cite{chan2023missing}. Convolutional Neural Networks (CNNs) have also been directly applied for this purpose, as shown by \cite{zhuang2019innovative}, whose model demonstrated better performance compared to the state of the art.

\section{Methodology} \label{Section: Methodology}

In the evolving landscape of transportation and human mobility forecasting, LLMs have become critical tools, offering innovative perspectives and methodologies for analyzing various complex datasets such as sensor datasets, map datasets, traffic flow datasets and route datasets, among others. This section dives deep into the various kinds of approaches to leverage LLMs within this domain, categorizing these approaches into two distinct sets of techniques: Processing (Tokenization, Prompt, Embedding) and Model Framework (Fine-tune, Zero-Shot/Few-Shot, Integration). As illustrated in Figure 2, which provides a general pipeline of LLM application for time series forecasting in transportation systems, the Processing techniques help users to create more LLM-friendly input data and manipulate LLM output data in various ways. The Model Framework section focuses on unblocking more potentials of LLMs for making more accurate predictions. Specifically, for Fine-tune and Zero-Shot/Few-Shot, we explore how to refine LLMs, while for Integration, we investigate better ways to fit LLMs into larger frameworks—considered an optional step in the overall pipeline. The final output of the pipeline can vary, including imputation, traffic chatbot, prediction and more. 

\begin{figure*}[!t]
\centering
\includegraphics[width=0.9\columnwidth]{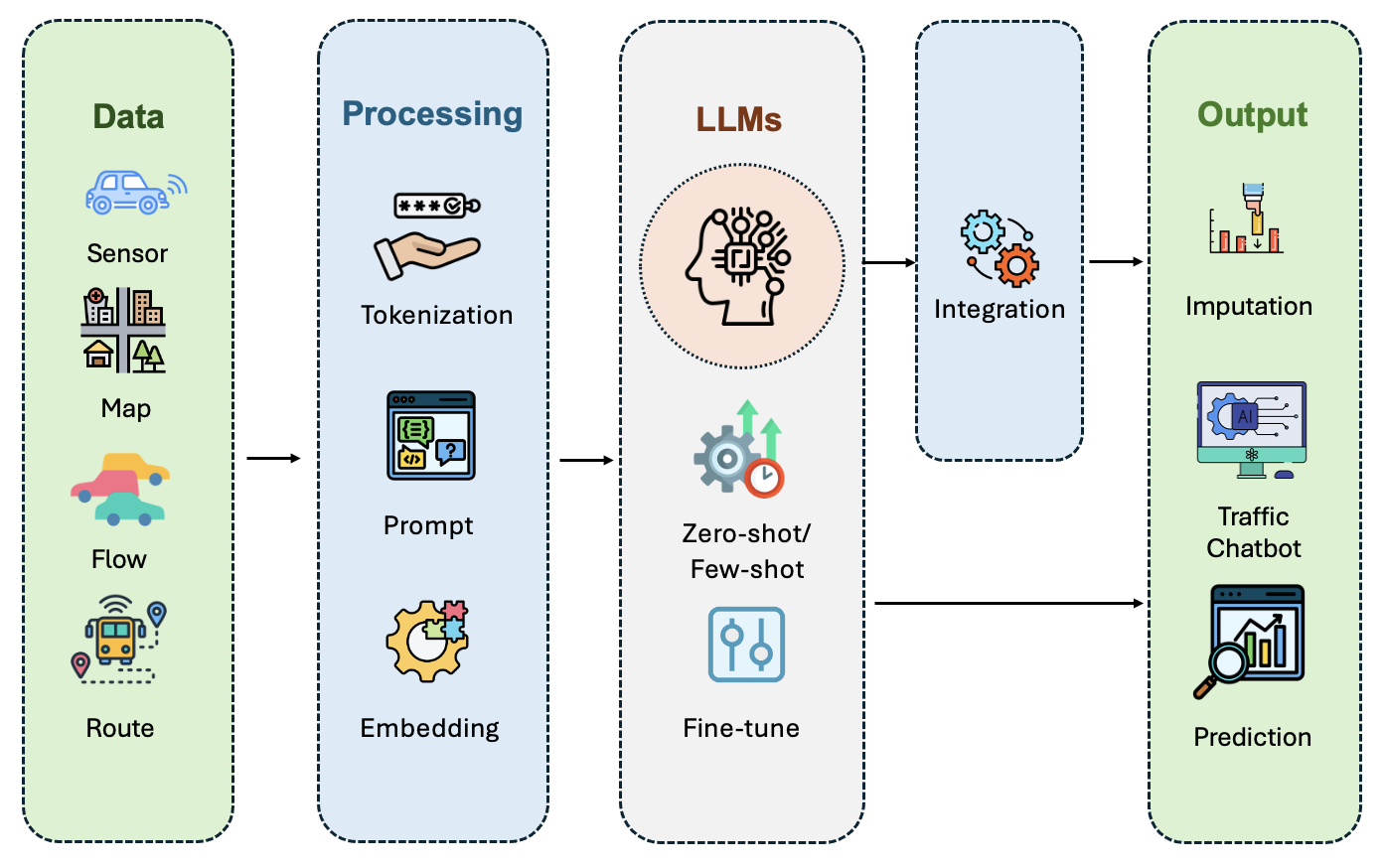}
\caption{Overview of methodologies in LLM pipeline for time series forecasting in Transportation Systems}
\vspace{-0.6cm}
\label{fig:2}
\end{figure*} 

Each of the techniques presents a unique way to interact with or utilize LLMs. From the processing perspective, \textbf{Tokenization} means introducing innovative tokenization technique for specific application scenarios (Section~\ref{Subsubsection: Token}). And \textbf{Prompt} refers to performing prompt engineering to provide more contexts and instructions to LLMs for better outcomes (Section~\ref{Subsubsection: Prompt}). \textbf{Embedding} denotes utilizing LLMs as encoders that generate meaningful deep representations from original data for the downstream processes (Section~\ref{Subsubsection: Embedding}). From the model perspective, \textbf{Fine-tune} means dedicated fine-tuning processes which tailor models to specific forecasting tasks (Section~\ref{Subsubsection: Fine-tune}). Then, \textbf{Zero-Shot/Few-Shot} refers to directly querying pretrained LLMs without any examples or with several concrete examples, respectively, while not modifying the LLMs' parameter weights (Section~\ref{Subsubsection: Zero-Shot/Few-Shot}). Finally, \textbf{Integration} denotes LLMs serving as an integral part of a larger infrastructure or pipeline (Section~\ref{Subsubsection: Integration}). By illustrating these techniques, we aim to provide a comprehensive understanding of how LLMs can be effectively deployed to enhance mobility prediction modeling in transportation systems.

\subsection{Data Processing}



\subsubsection{Tokenization} \label{Subsubsection: Token}

Tokenization is the process of breaking down raw textual data into a series of tokens as the input for querying LLMs, which makes data much more understandable and easier to analyze for LLMs.

Some tokenizers can be utilized in a notably straightforward way by directly breaking down the textual data into tokens, which is conceptually simple but still fairly effective, especially for time series data due to the limited informational breadth. But tokenization could also be leveraged in more sophisticated ways. For example, Liu et al. \cite{liu2024spatial} utilized a novel tokenizer, which defines timestamps at given locations as a token, then embeds tokens by a spatial-temporal embedding layer. After that, the authors performed embedding fusion to generate inputs for a partially frozen attention (PFA) LLM.

Through tokenization, researchers can transform different types of traffic data into tokens which can be easily consumed by LLMs \cite{gruver2024large}. While built-in tokenizers (e.g., Python NLTK) might usually be too generic, scientists can design tokenizers for specific applications. An appropriate tokenizer can be a great component to enhance the overall performance of LLM applications, especially for sophisticated sources like mobility data in transportation systems.

There is a typical example of tokenization technique, AuxMobLCast, proposed by Xue et al. \cite{xue2022leveraging}, which uses pre-trained language encoders (e.g., BERT, RoBERTa, and XLNet) to encode the raw mobility prompts into two sets of tokens, (1) contextual tokens, carrying the contextual information such as temporal data, and (2) numerical tokens, containing the numerical human mobility information such as number of visits to a place-of-interest (POI). These two sets of tokens are ready to be learned simultaneously by the transformer-based decoder (e.g., GPT-2) later. In addition, the authors introduce the [CLS] token in the initial prompt and take the feature embedding of this special token as the input for a fully connected layer followed by a softmax layer, which empowers the framework to be able to perform the POI category classification.


\subsubsection{Prompt} \label{Subsubsection: Prompt}

Prompt, or prompt engineering, means the process of structuring inputs to LLMs by providing more contexts and instructions in addition to original queries.

In the domain of LLM applications in transportation forecasting, prompt engineering can play an important role. For instance, Lai et al. \cite{lai2023large} proposed LLMLight, a novel framework employing LLMs as decision-making agents for traffic signal control, which instructs LLMs with knowledgeable prompts containing real-time traffic conditions. Moreover, Xue et al. \cite{xue2024prompt} introduced an innovative prompt mining framework in language-based human mobility forecasting, including a prompt generation stage based on the information entropy of prompts and a prompt refinement stage to integrate mechanisms such as the chain of thought.

Prompt engineering allows for the exploitation of LLMs' vast knowledge bases and sophisticated understanding of spatio-temporal mobility patterns without the need for computationally intensive training processes, which makes it a great way for researchers to explicitly guide LLMs. With more contexts and instructions in prompts, LLMs can better understand the tasks assigned by researchers and generate outputs following the expected response formatting \cite{xue2023promptcast}.

In \cite{xue2022leveraging}, the mobility prompting introduced by the authors can transform numerical temporal sequences into natural language sentences allowing the existing language models of intelligent digital agents (e.g., Alexa and Siri) to be leveraged directly. Prompt engineering resolves a major drawback of the numerical model paradigms, which mainly focus on extracting and modeling structured numeric data and are less effective in dealing with other formats of data. Additionally, pre-trained language models are great for understanding and interpreting the principal component analysis (PCA) loadings (a compact representation of significant patterns in the traffic dataset) through prompt engineering \cite{de2023llm}. LLMs' (e.g., LLaMA-2, GPT-4, Zephyr-7b-$\alpha$) prompts proposed by the authors were able to extract human-understandable patterns and relationships embedded within the traffic data by clearly instructing the LLMs for specific traffic accident forecasting tasks and feeding in the PCA loadings.





\subsubsection{Embedding} \label{Subsubsection: Embedding}

Embedding is the strategy of utilizing LLMs as encoding models, which produce meaningful deep representations (i.e., embeddings) of input queries, instead of textual/numerical results, as outputs. The output embeddings are then leveraged as inputs for downstream procedures in the framework.

There are various applications for the embedding strategy in transportation research. In \cite{xue2022leveraging}, a pipeline for predicting Place-of-Interest (POI) customer flows is proposed, which utilizes LLMs (e.g., BERT) as the encoder to produce feature embeddings for contextual and numerical tokens. Furthermore, LLMs can also be integrated into multimodal intelligent traffic systems by embedding text-based traffic information into feature vectors \cite{zheng2023chatgpt}.

Through embeddings, LLMs can serve as robust and effective encoders, which can capture key information from textual traffic data and convert them into the desired formatting for downstream deep learning networks. Furthermore, the mobility data in transportation systems usually contains multiple sources of information, including texts, images, audio \cite{chen2022end}, and so on. And the embedding technique is a straightforward way for LLMs' integration into sophisticated multimodal mobility forecasting frameworks.


The functionality of language models in Graph Transformer-based Traffic Data Imputation (GT-TDI) is to serve as the information extractors from semantic descriptions of historical traffic data, and the language models will output embedded semantic tensors \cite{zhang2024semantic}. Together with geographic edges, pattern edges, and incomplete traffic data, the semantic embedding from language models empowers GT-TDI's ability to impute missing traffic data effectively.

\subsection{Model Framework}
\subsubsection{Fine-tune} \label{Subsubsection: Fine-tune}
Fine-tuning is the process of feeding a dataset containing task-specific examples to update the weights of parameters in pre-trained LLMs through back-propagation.

Fine-tuned LLMs can effectively work with time series data, including mobility information in transportation systems. For instance, LLM4TS, an LLM-powered time series prediction framework, uses fine-tuned GPT-2 as its backbone model, which has good capability in interpreting temporal data \cite{chang2023llm4ts}. Moreover, Liu et al. \cite{liu2024can} proposed STG-LLM, an innovative approach for spatial-temporal forecasting, which also leverages GPT-2 by fine-tuning a small number of its parameters to enable its understanding of the semantics of researcher-defined spatial-temporal tokens.

Fine-tuning can provide researchers with a customized LLM which can be more accurate and effective for a given application domain with relatively low costs \cite{alt2023domain}. Also, fine-tuned LLMs can usually better understand inputs from researcher-designed tokenizers and be more likely to produce outputs in needed formats. Therefore, fine-tuning is a great approach to enhance the overall performance of small or generic LLMs in time series forecasting in transportation systems.

In GT-TDI, Zhang et al. \cite{zhang2024semantic} fine-tuned the pre-trained language models with task-specific data (e.g., spatiotemporal semantic descriptions) to align them with the distribution of datasets in the traffic domain. With fine-tuned parameters, the language models are more capable of imputing incomplete traffic data. Similarly in AuxMobLCast \cite{xue2022leveraging}, LLMs are fine-tuned for both the sequence generation and auxiliary category classification tasks and the joint training enhanced the proposed framework's capability to perform human mobility forecasting.



\subsubsection{Zero-Shot/Few-Shot} \label{Subsubsection: Zero-Shot/Few-Shot}

Zero-shot and few-shot learning are directly querying LLMs without updating their pre-trained parameters. The zero-shot technique only uses instructions in its prompts, while the few-shot technique contains several concrete examples in its prompts.

There are various zero-shot and few-shot applications in the domain of transportation systems. For instance, Li et al. \cite{li2024urbangpt} introduced UrbanGPT, an urban traffic spatio-temporal prediction framework, which also utilizes LLMs' zero-shot reasoning. Furthermore, few-shot prompts can provide LLMs with more traffic domain knowledge contained in text descriptions, so that LLMs can better consider spatial-temporal factors and their inter-dependencies in traffic prediction tasks \cite{guo2024explainable}.


Modern LLMs have demonstrated strong performance at tasks defined on-the-fly without fine-tuning \cite{brown2020language}. The zero-shot technique can achieve great task-agnostic performance, while the few-shot technique can produce even better outcomes. Without the need for training, LLMs can already be a great ingredient in the development of time series prediction frameworks in transportation systems.

Without further fine-tuning or training, TrafficGPT directly leverages pre-trained language models (e.g., GPT-3.5, ChatGLM3-6B, Qwen-14B-Chat, and InternLM-Chat-20B) to perform deductive reasoning, facilitated by the orchestration of the task request, the set of available traffic foundation models (TFMs), and the reasoning history in the prompts \cite{zhang2024trafficgpt}. In addition, de Zarz{\`a} et al. \cite{de2023llm} also proposed to apply the pre-trained LLMs without training. With detailed instructions, well-structured prompts, and contextual information (e.g., PCA loading strings, and the type of traffic accidents), the LLMs were able to perform extraction of human-understandable patterns and relationships embedded within the traffic data through zero-shot inferences.

\subsubsection{Integration} \label{Subsubsection: Integration}

Integration means that an LLM serves as an integral component to process or produce informative intermediate results in a large framework.

LLM integration has been widely applied in the field of time series and spatio-temporal forecasting including traffic forecasting \cite{zhang2024large}. For example, Ren et al. \cite{ren2024tpllm} proposed TPLLM, a traffic prediction framework which leverages GPT-2 as the base LLM to provide embedding inputs for downstream tasks, including traffic flow prediction, and traffic missing data imputation. Also, in ST-LLM, a framework introduced by Liu et al. \cite{liu2024spatial}, a PFA LLM is utilized for training on traffic feature datasets and inferring on new data to produce intermediate results for the downstream regression task to perform spatial-temporal prediction.

On the one hand, through integration with different types of models (e.g., computer vision, speech, etc.), LLMs can be leveraged effectively in multimodal forecasting tasks. On the other hand, LLMs' capabilities could be augmented through integration, because LLMs can encode textual traffic data into insightful embeddings which can be easily consumed by other deep learning models.

In TrafficGPT, Zhang et al. \cite{zhang2024trafficgpt} enabled iterative interactions between LLMs and the necessary TFMs to enhance LLMs' understanding of operational contexts within the traffic domain. This integration allows TrafficGPT to leverage multi-modal data as a source, providing more comprehensive support for various traffic tasks—a capability that cannot be achieved by either LLMs or TFMs alone. There are also more straightforward integrations of LLMs with other networks. A refined version of BERT, called TrafficBERT, was proposed by Jin et al. \cite{jin2021trafficbert}, and it has the ability to encode continuous traffic sequence data by taking linearly transformed inputs through stacks of transformer encoders. In the end, TrafficBERT is integrated with the final linear layer to generate predicted traffic sequences.

\begin{table*}[]
    \caption{Taxonomy of LLM Applications in Forecasting Tasks in Transportation and Urban Systems}
    \label{tab:Taxonomy}
    \centering
    \begin{threeparttable}
    \scalebox{0.91}{
    \begin{tabular}{p{3.5cm} p{2.2cm} p{0.5cm} p{0.5cm} p{0.5cm} p{0.5cm} p{0.5cm} p{0.5cm} p{2.5cm} p{1.4cm}}
        \hline
        & \\[-1.5ex]
        \multirow{2}{*}{Method} & \multirow{2}{*}{Domain} & \multicolumn{3}{c}{Data Processing} & \multicolumn{3}{c}{Model Framework} & \multirow{2}{*}{\makecell{Pretrained\\ Model}} & \multirow{2}{*}{Code}\\
        \hhline{~~|---||---|~}
        & \\[-2ex]
        & & T & P & E
        & F &  Z/F & I  \\
        \hline
        \\[-2ex]

        de Zarz{\`a} et al. \cite{de2023llm} & Traffic & \ding{55} & \ding{51} & \ding{55} & \ding{51} & \ding{51} & \ding{51} & LLaMA2, GPT-4, Zephyr-7b-$\alpha$ & No \\ 
        \hline
        \\[-2ex]
        TF-LLM \cite{guo2024explainable} & Traffic & \ding{55} & \ding{51} & \ding{55} & \ding{51} & \ding{51} & \ding{55} & LLaMA2 & No \\ 
        \hline
        \\[-2ex]
        MobilityGPT \cite{haydari2024mobilitygpt} & Human Mobility & \ding{51} & \ding{51} & \ding{55} & \ding{51} & \ding{55} & \ding{51} & From scratch & No \\ 
        \hline 
        \\[-2ex]
        STLLM \cite{zhang2023spatio} & Traffic & \ding{55} & \ding{51} & \ding{51} & \ding{55} & \ding{55} & \ding{51} & GPT-3.5 & No \\ 
        \hline 
        \\[-2ex]
        ST-LLM \cite{liu2024spatial} & Traffic, Demand & \ding{51} & \ding{55} & \ding{55} & \ding{51} & \ding{51} & \ding{51} & GPT-2, LLaMA2 & No \\ 
        \hline
        \\[-2ex]
        GT-TDI \cite{zhang2024semantic} & Imputation & \ding{55} & \ding{51} & \ding{51} & \ding{51} & \ding{55} & \ding{51} & BERT, GPT-3.5 & No \\ 
        \hline
        \\[-2ex]
        AuxMobLCast \cite{xue2022leveraging} & Human mobility & \ding{51} & \ding{51} & \ding{51} & \ding{51} & \ding{51} & \ding{51} & GPT-2 & Yes\textsuperscript{[1]} \\ 
        \hline
        \\[-2ex]
        Zheng et al. \cite{zheng2023chatgpt} & Demand & \ding{55} & \ding{51} & \ding{55} & \ding{51} & \ding{51} & \ding{51} & ChatGPT & No \\ 
        \hline
        \\[-2ex]
        STG-LLM \cite{liu2024can} & Traffic & \ding{51} & \ding{51} & \ding{51} & \ding{51} & \ding{51} & \ding{51} & GPT-2 & No \\ 
        \hline
        \\[-2ex]
        TrafficGPT \cite{zhang2024trafficgpt} & Demand  & \ding{55} & \ding{51} & \ding{51} & \ding{55} & \ding{51} & \ding{51} & GPT-3.5-turbo & Yes\textsuperscript{[2]} \\
        \hline
        \\[-2ex]
        UrbanGPT \cite{li2024urbangpt} & Traffic & \ding{51} & \ding{51} & \ding{55} & \ding{55} & \ding{51} & \ding{51} & Vicuna & Yes\textsuperscript{[3]} \\ 
        \hline
        \\[-2ex]
        TPLLM \cite{ren2024tpllm} & Traffic & \ding{51} & \ding{55} & \ding{51} & \ding{51} & \ding{55} & \ding{51} & GPT-2 & No \\ 
        \hline
        \\[-2ex]
        TrafficBERT \cite{jin2021trafficbert} & Traffic & \ding{55} & \ding{55} & \ding{55}  & \ding{51} & \ding{55} & \ding{51} & From scratch & No \\ 
        \hline

        Mo et al. \cite{mo2023large} & Human Mobility, Demand  & \ding{55} & \ding{51} & \ding{55} & \ding{55} & \ding{51} & \ding{55} & GPT-3.5 & No \\
        \hline
        \\[-2ex]
        UniST \cite{yuan2024unist} & Traffic & \ding{51} & \ding{51} & \ding{51} & \ding{51} & \ding{51} & \ding{55} & From scratch & No \\ 
        \hline
        \\[-2ex]
        GATGPT \cite{chen2023gatgpt} & Imputation & \ding{51} & \ding{55} & \ding{55} & \ding{51} & \ding{55}  & \ding{51} & GPT-2 & No \\ 
        \hline
        \\[-2ex]
        LLM-MPE \cite{liang2024exploring} & Human Mobility & \ding{51} & \ding{51} & \ding{55} & \ding{55} & \ding{51} & \ding{51} & GPT-4 & No \\ 
        \hline
        \\[-2ex]

        LLM-Mob \cite{wang2023would} & Human Mobility & \ding{55} & \ding{51} & \ding{55} & \ding{55} & \ding{51} & \ding{55} & GPT-3.5  & Yes\textsuperscript{[4]} \\ 
        \hline
        \\[-2ex]

        CPPBTR \cite{duan2019pre} & Traffic & \ding{55} & \ding{55} & \ding{55} & \ding{51} & \ding{55} & \ding{51} & From scratch & No \\ 
        \hline
        \\[-2ex]

        TFM \cite{wang2023building} & Traffic & \ding{55} & \ding{55} & \ding{55} & \ding{55} & \ding{55} & \ding{55} & From scratch & Yes\textsuperscript{[5]} \\ 
        \hline
        \\[-2ex]

    \end{tabular}
    }
    {
      \scriptsize
      \begin{tablenotes}
        \item[]
        Note: \textbf{T}okenization, \textbf{P}rompt, \textbf{E}mbedding, \textbf{F}ine-tune, \textbf{Z}ero-shot/\textbf{F}ew-shot, \textbf{I}ntegrate
        \item[]
        [1] https://github.com/cruiseresearchgroup/AuxMobLCast
        [2] https://github.com/lijlansg/trafficgpt 
        [3] https://github.com/HKUDS/UrbanGPT 
        \item[]
        [4] https://github.com/xlwang233/LLM-Mob 
        [5] https://github.com/SACLabs/TransWorldNG
      \end{tablenotes}
    }
   \end{threeparttable}
\end{table*}

\section{Applications} \label{Section: Applications}


In this section, we present recent innovative deep learning applications of LLMs and foundation models in the mobility analysis of transportation systems across various fields, including traffic forecasting, human mobility, demand forecasting, and missing data imputation. We have summarized the methods proposed in these research works in the taxonomy TABLE~\ref{tab:Taxonomy}.



\subsection{Traffic Forecasting}

Traditional statistical methods typically treat traffic forecasting as time series problems~\cite{wang2022novel, rahman2023deep}. One of the common approaches is to utilize autoregressive models (e.g., ARIMA) to predict time series. After the advent of deep learning, recurrent neural network (RNN)-based and knowledge-based methods were introduced in 
times series forecasting~\cite{wang2022novel, rahman2023deep, zhan2018small}. For instance, Ma et al. \cite{ma2015large} explored deep learning methods in this field through the combination of deep restricted Boltzmann machine with RNN to model and predict the evolution of traffic congestion. Using GPS data, this method achieves high prediction accuracy, providing valuable insights for congestion mitigation. Furthermore, focusing on scalability and efficiency, Monteil et al. \cite{monteil2021model} compared multiple deep learning models with simpler predictors for long-term, large-scale traffic predictions, emphasizing the importance of prediction accuracy, training time, and model size.

However, RNN-based methods are hard to learn long-term temporal dependencies, and it is difficult for domain knowledge-based methods to model temporal dependency automatically. A pioneering approach in this realm, Pre-trained Bidirectional Temporal Representation (PBTR), can overcome the limitations of these methods. PBTR utilizes the Transformer encoder to predict crowd flows in gridded regions and demonstrates exceptional capability in modeling long-term temporal dependencies within an encoder-decoder framework, significantly enhancing prediction accuracy \cite{duan2019pre}. Furthermore, the Traffic Transformer model demonstrates the potential of deep learning architectures in modeling time series and spatial dependencies in traffic forecasting, significantly outperforming traditional models \cite{cai2020traffic}.

Building on the achievements of previous deep learning models, the application of LLMs further underscores the potential of innovative approaches in the domain of traffic forecasting. For instance, TrafficBERT uses transformers for traffic flow prediction, outperforming traditional statistical and deep learning models. It efficiently utilizes large-scale traffic data and employs multi-head self-attention to navigate the complexities of various road conditions without necessitating road-specific or weather data \cite{jin2021trafficbert}. Moreover, the application of LLMs extends to the Generative Graph Transformer (GGT) model, designed for city-scale traffic forecasting. Treating traffic flow and interactions as sequences, GGT comprehends and predicts complex traffic patterns, facilitating more dynamic and accurate predictions of traffic conditions, thereby aiding in improved traffic management and planning \cite{wang2023building}.

Recent innovations in LLM application include STLLM, which integrates LLM with a mutual information maximization paradigm of cross-view to capture implicit spatio-temporal dependencies and preserve spatial semantics for traffic flow in urban areas \cite{zhang2023spatio}. In addition, Liu et al. \cite{liu2024can} proposed STG-LLM, which adapts LLMs for spatial-temporal forecasting through a spatial-temporal graph tokenizer and adapter, bridging the comprehension gap between complex spatial-temporal data and LLMs. Another study focuses on employing LLMs for forecasting traffic accidents, using the Large Language and Vision Assistant (LLaVA), a bridge between visual and linguistic information powered by the Visual Language Model (VLM), in conjunction with deep probabilistic reasoning to improve the real-time responsiveness of autonomous driving systems \cite{de2023llm}. Furthermore, Guo et al. \cite{guo2024explainable} proposed TF-LLM, an innovative approach to generate interpretable traffic flow predictions, which leverages LLaMA2 to process multimodal traffic data, including system prompts, real-time spatial-temporal data, and external factors to make predictions and provide explanations about traffic flow. Finally, Ren et al. \cite{ren2024tpllm} introduced TPLLM, a traffic prediction framework based on pretrained LLMs, which demonstrates the efficacy of combining LLMs with convolutional and graph convolutional networks for traffic prediction, especially in scenarios with limited historical data.

Collectively, these studies underscore the transformative potential of deep learning and LLMs in traffic forecasting, offering innovative solutions for managing and understanding complex transportation systems.




\subsection{Human Mobility}


LLMs have become pivotal tools in contemporary research aiming to understand and forecast the complexities of human mobility dynamics, surpassing traditional models. Wang et al. \cite{wang2023would} introduced LLM-Mob, a novel method using LLMs for accurate and interpretable human mobility prediction by leveraging language understanding and reasoning capabilities, along with new concepts which capture both short-term and long-term human movement dependencies and context-inclusive prompts to improve the accuracy of predictions. Additionally, LLMs can be integrated to forecast human mobility and visitor flows to POI by utilizing all kinds of information, such as numerical values and contextual semantic information, as components in natural language inputs \cite{xue2022leveraging}. Furthermore, LLM-MPM, a framework for human mobility prediction under public events, shows the unprecedented ability of LLMs to process textual data, learn from minimal examples, and generate human-readable explanations \cite{liang2024exploring}.

In addition to the direct application of LLMs on human mobility prediction, researchers have also introduced generative models inspired by LLMs. For instance, Haydari et al. \cite{haydari2024mobilitygpt} proposed a geospatially-aware generative model, MobilityGPT, to capture human mobility characteristics and generate synthetic trajectories. Leveraging a gravity-based sampling method to train a transformer for semantic sequence similarity, MobilityGPT can ensure its controllable generation of semantically realistic geospatial mobility data to reflect real-world characteristics.

\subsection{Demand Forecasting}

Numerous LLM applications have been proposed in the domain of demand forecasting. For example, Liu et al. \cite{liu2024spatial} introduced the Spatial-Temporal Large Language Model (ST-LLM) designed for traffic demand prediction, incorporating a spatial-temporal embedding module to learn the spatial locations and global temporal representations of tokens before embedding fusion and feeding into LLMs. ST-LLM can effectively predict taxi and bike demands to enable efficient allocation and scheduling of vehicles. Moreover, Mo et al. \cite{mo2023large} highlighted a shift toward utilizing LLMs' reasoning abilities for complex predictions in travel demand and behavior studies without traditional data-based training. By carefully crafting prompts with travel characteristics, individual attributes, and domain knowledge, the study demonstrates that LLMs can predict travel choices accurately and provide logical explanations for the predictions. Tested against standard models, such as multinomial logit and random forests, the LLM approach shows competitive accuracy and F1-score. 

Inspired by general LLMs, there are also domain-specific LLMs trained from scratch in traffic studies. Yuan et al. \cite{yuan2024unist} introduced UniST, a universal model for urban spatio-temporal prediction, addressing the need for a versatile model capable of adapting to various urban scenarios with different spatio-temporal features. UniST leverages elaborate masking strategies for generative pre-training and employs spatio-temporal knowledge-guided prompts to align and utilize shared knowledge across different scenarios effectively. This approach enables UniST to perform well in diverse prediction tasks, including demand forecasting, demonstrating its universality and effectiveness through extensive experiments across multiple cities and domains, notably excelling in few-shot and zero-shot settings.

\subsection{Missing Data Imputation}


Several studies represent how LLMs help traffic spatial-temporal imputation tasks. In \cite{zhang2024semantic}, GPT-3.5 is applied to generate human-like texts to fine-tune a BERT-based text model, which generates traffic semantic tensors from the semantic descriptions. This method enhances the accuracy of filling in missing and updating inaccurate traffic data, demonstrating the capability of LLMs in interpreting complex spatial-temporal traffic patterns. Another study, GATGPT by Chen et al. \cite{chen2023gatgpt}, also claims its effectiveness in spatial-temporal imputation tasks, which leverages pre-trained LLMs with a graph attention network for spatial-temporal imputation. This method is designed to efficiently handle missing data in multivariate time series by capturing both spatial and temporal dependencies.

\section{Challenges and Outlook} \label{Section: Outlook}









In this section, we discuss the limitations of current research and potential further research directions in the field of mobility forecasting in contemporary transportation systems with LLMs.

\subsection{Interpretability and Explainability}

Unlike traditional computer science or artificial intelligence research, the studies in traffic forecasting do care about integrating transportation science and how transportation domain knowledge helps forecasting models generate interpretable results. Therefore, it is crucial not only to make accurate forecasts for transportation systems but also to understand why a model made a particular forecast or decision, to better arrange traffic and understand human mobility patterns.

Besides the strong performance in forecasting tasks, LLMs are also new paradigms that provide great interpretability and explainability. Interpretability means that LLMs can conveniently infer causalities while producing forecasting results \cite{creswellselection}. Explainability refers that LLMs can generate and showcase human-like thought processes in natural language, such as Chain of Thought (CoT) \cite{wei2022chain, zhangautomatic, mitra2024compositional}. LLMs are suitable for both improving the performance of forecasting models and facilitating the interpretability and explainability for forecasting results in transportation domain.

However, at this point, most existing papers only present extensive experiments to demonstrate the effectiveness of the proposed methods, but ignore the interpretations of results and the explanations of the thought processes \cite{xue2022leveraging, jin2021trafficbert}. This practice not only underutilizes the unique ability of LLMs, but also makes it difficult for researchers to understand the incentives and rationales for LLMs behind LLMs producing certain results.


Also currently, many LLM-powered frameworks did not integrate with domain knowledge in transportation very well. For example, Ren et al. \cite{ren2024tpllm} introduced TPLLM, an LLM-based traffic prediction framework, which leverages the sequential nature of traffic data, similar to that of language. However, TPLLM does not incorporate established transportation-specific theories or models, such as traffic flow theory \cite{hoogendoorn2012traffic} and traffic assignment models \cite{patriksson2015traffic}, which might be crucial for more interpretable traffic predictions. And in \cite{jin2021trafficbert}, TrafficBERT, a BERT model pre-trained with large-scale traffic data, is proposed to forecast traffic flow on various types of roads. But TrafficBERT mainly treats traffic data as general spatio-temporal time series information, and does not include much transportation-specific background.

Therefore, a promising future direction for research in the transportation forecasting domain is utilizing LLMs to build interpretable and explainable modules with more emphasis on transportation domain knowledge. Such modules can be very beneficial for analyzing the inference results and diagnosing errors or unexpected behaviors. Also, the interpretability and explainability of LLMs will make the overall framework much more transparent by providing human-understandable rationales.

\subsection{Privacy Concerns about LLM-Powered Transportation Frameworks}


Privacy is a key bottleneck for collecting real-world data in transportation systems \cite{fries2012meeting}, and it is also a major concern for wider utilization of LLMs \cite{pan2020privacy}. Therefore, even though the strong generalization ability is an important advantage of LLMs \cite{xu2021raise}, in the transportation domain, researchers may still face obstacles due to the lack of publicly available datasets suitable for fine-tuning general-purpose LLMs into transportation-specific LLMs.

First, it is challenging to protect data privacy in intelligent transportation (ITS) devices, which are crucial for collecting transportation data. This is because, to ensure data security and integrity, ITS devices rely on secret keys \cite{munir2020design}. However, many ITS devices lack the capability or resources to securely store and manage secret keys generated for secure communication or data transfer, making the privacy of collected traveler data vulnerable \cite{hahn2019security}.

Second, LLMs may leak private information and compromise data privacy \cite{pan2020privacy}. One reason is that LLMs are memorizing training data, and it has been proven that extracting sensitive information from them is a practical threat \cite{carlini2021extracting}. Also, LLMs may have the ability correctly infer private information, meaning that even if users only provide publicly available data, LLMs can still sometimes infer and disclose users' correct private information \cite{weidinger2021ethical}. Research also suggests that LLMs should be trained only on data explicitly produced for public use \cite{brown2022does}.

Third, intensively interactions with LLMs and providing private data to query LLMs in transportation systems make it even more difficult to maintain data privacy. Many LLM-powered transportation models require private or sensitive information, including but not limited to real-time traffic flow videos, mobility data, temporal information, vehicle sensory data, and even conversational data from nearby vehicles \cite{de2023llm, xue2022leveraging, zhang2024trafficgpt}.


Recently, advancements have been made in maintaining data privacy for LLMs and forecasting frameworks in transportation systems, particularly through the use of differential privacy. For instance, an efficient differentially private stochastic gradient descent mechanism was proposed, which can be applied to fine-tune LLMs and has theoretical privacy guarantees \cite{dupuy2022efficient}. Zhang et al. \cite{zhang2021fastgnn} introduced a privacy-preserving federated learning approach for traffic speed forecasting, utilizing a differential privacy-based adjacency matrix to protect topological information. In addition, another privacy-preserving blockchain-based framework for traffic flow prediction has been proposed, which stores model updates from distributed vehicles on the blockchain and leverages a differential privacy method with a noise-adding mechanism to enhance location privacy protection \cite{qi2021privacy}. Furthermore, LLM agents can also incorporate homomorphic encryption schemes and attribute shuffling mechanisms to safeguard user privacy \cite{zhang2024privacyasst}. Finally, many of the transportation companies have been using databases and cloud planforms from big techs like Oracle and Microsoft, there have been mature solutions for protecting data privacy and security between the companies, the LLM application could be developed based on those mature solutions.


\subsection{Cost and Legality}

Application development based on LLMs is significantly more costly than traditional model development. As the most advanced AI technique, LLM development requires substantially more computational resources compared to traditional models \cite{patil2024review}, like hundreds or thousands of GPUs/TPUs as computational resources. The financial cost for LLM application development and maintenance will be very expensive. Hybrid professionals in transportation and AI are also required for LLM application development. Unfortunately, transportation professionals are probably not proficient in LLM development, the companies will need to invest heavily in hiring new qualified staff, and training current staff to learn to use LLMs. To address these challenges, traditional transportation companies may have to work closely with IT giants that have been developing LLMs to access computational resources and cooperate with their researchers to develop the applications together.

Another issue is that the transportation data sources in industry may be not sufficient for application development. Many important data sources, such as the environmental and traffic condition data, are collected by roadside cameras and sensors which are installed by government institutions, and traffic data collection based on that remains one of the biggest challenges for intelligent transportation systems even though the governments have been investing much money on that \cite{guerrero2018sensor}. Therefore, it is important for the companies to collaborate with the government to gain the necessary data, and this collaboration necessitates stringent data privacy and security measures to protect sensitive information. Companies must navigate these legal frameworks to ensure that data usage does not infringe on individual privacy rights. Furthermore, there are technical challenges related to data interoperability and standardization. Different government agencies and companies may use various formats and protocols for data collection and storage, making it difficult to aggregate and analyze the data efficiently. Establishing common standards and protocols is essential for seamless data integration and utilization.

\subsection{Insufficient Open Data Resources}

Despite the importance of open data in transportation research, the availability of datasets in this field remains quite limited. Although most studies listed in TABLE 1 utilize publicly available datasets, these datasets are primarily confined to specific geographical areas such as California (PeMS managed by Caltrans), New York City, and Chicago \cite{varaiya2009freeway, li2024urbangpt, yuan2024unist}. Alternative sources, such as the Beijing taxi trajectories utilized by \cite{duan2019pre}, require significant data preprocessing, while the SUMO dataset employed by \cite{wang2023building} is suitable only for highly specific tasks. Furthermore, certain datasets, such as the Foursquare New York City (FSQ-NYC) dataset referenced by \cite{wang2023would}, are no longer accessible due to inactive download links. The current situation in limited available transportation datasets underscores that current research is predominantly concentrated on a few locations, leaving much of the world without accessible traffic data.

Another challenge is that most of the existing datasets consist of mainly numeric data and lack textual data, limiting their compatibility with LLMs. The few datasets containing free text are typically collected for specific research purposes. For instance, the Fatality Analysis Reporting System (FARS) dataset referenced by \cite{de2023llm} is primarily used for traffic accident analyses, while the Barclays Center event data collected by \cite{liang2024exploring} was specifically scraped from the official website for a focused case study and is not part of a standardized database that could be utilized for other studies.

There is a critical need to develop more open-source datasets for transportation forecasting with consistent standards across regions. This would enable LLMs trained on data from one location, such as NYC, to be readily applied to other cities like Philadelphia or Boston. Public databases should also be updated regularly to ensure that the available resources remain functional. Furthermore, to enhance the effectiveness of LLMs in this domain, datasets should incorporate more associated multi-modal data or retain embedded original free text content, allowing for more versatile and in-depth analysis across diverse tasks and applications in transportation systems.

\section{Conclusion and Future Work} \label{Section: Conclusion}

We present a comprehensive and up-to-date study of LLMs and their variants tailored for the analysis of forecasting problems in transportation and human mobility scenarios. By introducing a new taxonomy, we categorize and assess prominent techniques in each domain, highlighting their respective strengths, limitations, and practical applications. We aim to not only describe the current landscape but also provide a structured perspective that could serve as a foundational reference for future work in this emerging field.


Looking forward, we see numerous research opportunities to advance the use of LLMs in forecasting tasks in transportation systems. Key areas include the development of interpretable models that integrate theories in transportation domain, the establishment of privacy-preserving techniques suitable for LLMs in real-world deployments, and the creation of standardized, open-source datasets that support cross-regional transportation applications. We aspire for this survey to act as a spark, igniting further interest and sustaining a deep-seated enthusiasm for research in LLMs and their uses in transportation systems.


\section{AUTHOR CONTRIBUTIONS} \label{Section: AUTHOR CONTRIBUTIONS}
The authors confirm contribution to the paper as follows: study conception and design: Zijian Zhang, Yujie Sun, Zepu Wang; data collection: Zijian Zhang, Yujie Sun, Zepu Wang; analysis and interpretation of results: Zijian Zhang, Yujie Sun, Zepu Wang, Yuqi Nie, Xiaobo Ma, Ruolin Li, Peng Sun; draft manuscript preparation: Zijian Zhang, Yujie Sun, Zepu Wang, Yuqi Nie, Xiaobo Ma, Ruolin Li, Peng Sun, Xuegang Ban. All authors reviewed the results and approved the final version of the manuscript.

\section{Declaration of Conflicting Interests}
The authors declared no potential conflicts of interest with respect to the research, authorship, and/or publication of this article.

\section{Funding}
This study is supported by NSFC Grant (62250410368).

\newpage

\bibliographystyle{trb}
\bibliography{citation}
\end{document}